\documentclass[sigconf]{acmart}

\usepackage{url}
\usepackage{graphicx}
\usepackage{amsmath}
\usepackage{amsthm}
\usepackage{booktabs}
\usepackage{algorithm}
\usepackage{algorithmic}
\usepackage{booktabs}
\usepackage{multirow}
\usepackage[utf8]{inputenc}

\newtheorem{defn}{Definition}

\AtBeginDocument{%
  \providecommand\BibTeX{{%
    \normalfont B\kern-0.5em{\scshape i\kern-0.25em b}\kern-0.8em\TeX}}}

\begin{document}


\title{PFA: Privacy-preserving Federated Adaptation for Effective Model Personalization}

\author{Bingyan Liu}
\email{lby\_cs@pku.edu.cn}
\affiliation{%
  \institution{MOE Key Lab of HCST, Dept of Computer Science, School of EECS, Peking University}
  \city{Beijing}
  \state{China}
  \postcode{100871}
}

\author{Yao Guo}
\authornote{Yao Guo is the corresponding author.}
\email{yaoguo@pku.edu.cn}
\affiliation{%
  \institution{MOE Key Lab of HCST, Dept of Computer Science, School of EECS, Peking University}
  \city{Beijing}
  \state{China}
  \postcode{100871}
}

\author{Xiangqun Chen}
\email{cherry@pku.edu.cn}
\affiliation{%
  \institution{MOE Key Lab of HCST, Dept of Computer Science, School of EECS, Peking University}
  \city{Beijing}
  \state{China}
  \postcode{100871}
}







\renewcommand{\shortauthors}{Liu et al.}

\begin{abstract}
Federated learning (FL) has become a prevalent distributed machine learning paradigm with improved privacy. After learning, the resulting federated model should be further personalized to each different client. While several methods have been proposed to achieve personalization, they are typically limited to a single local device, which may incur bias or overfitting since data in a single device is extremely limited. In this paper, we attempt to realize personalization beyond a single client. The \textit{motivation} is that during the FL process, there may exist many clients with similar data distribution, and thus the personalization performance could be significantly boosted if these similar clients can cooperate with each other. Inspired by this, this paper introduces a new concept called \textit{federated adaptation}, targeting at adapting the trained model in a federated manner to achieve better personalization results. However, the key challenge for federated adaptation is that we could not outsource any raw data from the client during adaptation, due to \textit{privacy} concerns. In this paper, we propose \textbf{PFA}, a framework to accomplish \textbf{P}rivacy-preserving \textbf{F}ederated \textbf{A}daptation. PFA leverages the sparsity property of neural networks to generate privacy-preserving representations and uses them to efficiently identify clients with similar data distributions. Based on the grouping results, PFA conducts an FL process in a group-wise way on the federated model to accomplish the adaptation. For evaluation, we manually construct several practical FL datasets based on public datasets in order to simulate both the \textit{class-imbalance} and \textit{background-difference} conditions. Extensive experiments on these datasets and popular model architectures demonstrate the effectiveness of PFA, outperforming other state-of-the-art methods by a large margin while ensuring user privacy. We will release our code at: \textit{\url{https://github.com/lebyni/PFA}}.

\end{abstract}

\ccsdesc[500]{Human-centered computing~ Ubiquitous and mobile computing}
\ccsdesc[500]{Computing methodologies~Neural networks}
\ccsdesc[500]{Security and privacy~Privacy protections}

\keywords{Decentralized AI, Federated Learning, Neural Networks, Personalization, Privacy}


\maketitle

\section{Introduction}

\begin{figure}[t]
\centering
\includegraphics[width=0.8\columnwidth]{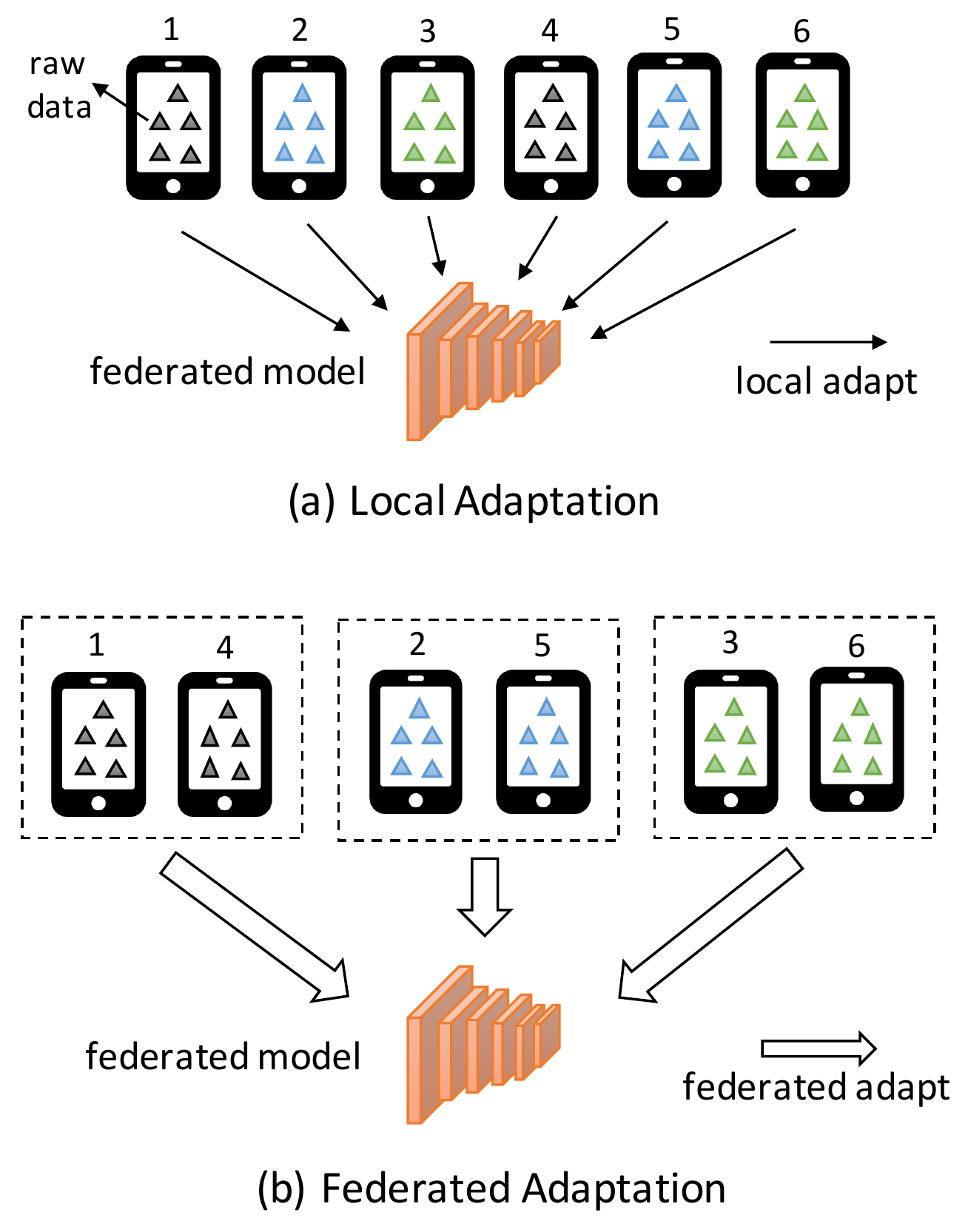} 
\caption{Comparison of two adaptation policies. Different from local adaptation that adapts the federated model with the data of each single client, the proposed federated adaptation approach aims to identify clients with similar data distributions and uses them to collaboratively adapt the federated model. Here the color of the triangle represents the data distribution.}
\label{fig:local_fed}
\end{figure}

Federated learning (FL) \cite{mcmahan2017communication,zhao2018federated} has been proposed and drawn great attention due to its capability to collaboratively train a shared global model under the decentralized data settings. A typical method to implement federated learning is the Federated Averaging (FedAvg) \cite{mcmahan2017communication}, which generates a global model by averaging the local parameters uploaded from each client. During the process, we do not exchange the sensitive raw data in each client and thus protect user privacy. In recent years, there have been extensive applications for deploying FL in practice, such as loan status prediction, health situation assessment, and next-word prediction \cite{hard2018federated,yang2018applied,yang2019federated}.

Although FL has been proven effective in generating a better federated model, it may not be the optimal solution for each client since the data distribution of clients is inherently non-IID (non-independently identically distribution). Here we believe the distribution not only refers to the statistical heterogeneity (e.g., the number of image category in a certain client) as prior work simulated \cite{mcmahan2017communication,li2019convergence}, but also includes the situation where the object is identical while the background is different (e.g. one user may take photos mostly indoors while another mostly outdoors). Under this condition, each client should own a personalized model rather than a global shared model in order to better fit its distinctive data distribution.

Fortunately, the research community has noticed the \textit{data heterogeneity} issue and several methods have been proposed to address the problem. For example, Wang \textit{et al.} \cite{wang2019federated} accomplished personalization by further fine-tuning the federated model with the local data in each client. Yu \textit{et al.} \cite{yu2020salvaging} extended the above work, where the federated model can be personalized via three schemes: fine-tuning, multi-task learning, and knowledge distillation. Although such methods can facilitate personalization to some extent, they have a significant drawback –  the personalization process is restricted in a single device, which may introduce some bias or overfitting problem since data in a device is extremely limited. Our \textbf{intuition} is that, in the FL process, there may exist many other clients that own similar data distribution to a certain client. If these clients can be aggregated and benefit from each other, the performance will definitely outperform the local adaptation schemes because more valuable data are utilized to personalize the federated model, mitigating the bias or overfitting problem as well as extending the useful knowledge.

Inspired by this, we introduce a new concept called \textbf{federated adaptation}, which is defined as \textit{adapting the federated model in a federated manner to achieve better personalization results.}  As shown in Figure \ref{fig:local_fed}, federated adaptation attempts to use the clients with similar distribution to collaboratively adapt the federated model, rather than just adapting it with the data in a single device. Compared to the traditional federated learning, federated adaptation has the following differences: (1) The adaptation objective is a federated trained model, which means that an FL process should be conducted first before the adaptation begins; (2) Instead of using the whole clients or randomly sampling them as the traditional FL does, in federated adaptation, the federated clients must be selective in order to guarantee the distribution matching. 


To the best of our knowledge, there are no existing work that conduct personalization in a federated manner. In order to accomplish federated adaptation, one key challenge is that the raw data in each client cannot be outsourced due to \textit{privacy} concerns. To solve this issue, this paper proposes \textbf{PFA}, a prototype framework for achieving personalization via \textbf{P}rivacy-preserving \textbf{F}ederated \textbf{A}daptation. The key idea behind PFA is that \textit{we can leverage the sparsity property of neural networks to generate a privacy-preserving representation which can then be used to replace the raw data for client grouping during the adaptation process}. Specifically, given a federated model, PFA first extracts the client-related sparsity vector as a privacy-preserving representation, and uploads them to the server to distinguish different distributions across clients (\textbf{Section \ref{sec:PFE}}). By employing Euclidean Distance to measure the similarity between these sparsity vectors, PFA is able to generate a matrix that describes the distribution similarity degree among clients (\textbf{Section \ref{sec:FSC}}). Based on the matrix, PFA manages to precisely group the clients with similar data distribution and conduct a group-wise FL process to accomplish the adaptation (\textbf{Section \ref{sec:CSM}}). 

Note that existing benchmark datasets of the non-IID setting are designed to simulate the \textit{class-imbalance} scenario \cite{mcmahan2017communication}, which is incomplete to demonstrate the real-world applications. Therefore, we construct several datasets based on some public datasets to simulate both the \textit{class-imbalance} and \textit{background-difference} conditions, in order to better represent the characteristics of practical FL environments (\textbf{Section \ref{sec:experiemnt_setup}}). 

We evaluate PFA on the constructed datasets and compare it to the FL baseline and three state-of-the-art local adaptation schemes for personalization. The results show that PFA outperforms other methods by up to 8.34\%, while ensuring user privacy. Besides, we conduct a number of detailed analyses (e.g., convergence analysis, privacy analysis) to further demonstrate the necessity and effectiveness of PFA.

To summarize, this paper makes the following main contributions:

\begin{itemize}

\item \textbf{A new idea to achieve better personalization in FL.} We introduce \textit{federated adaptation}, which conducts personalization in a federated manner rather than focusing on a local client.

\item \textbf{A novel approach to represent data.} We utilize the sparsity property of neural networks to represent the raw data in clients. This representation offers a privacy-preservation way to describe the client-related data distribution accurately.

\item \textbf{A comprehensive federated adaptation framework.} We propose PFA, a framework to personalize the federated model via privacy-preserving federated adaptation. To the best of our knowledge, this is the first attempt in the literature to explore and study the concept of federated adaptation.


\item \textbf{Detailed experiments to evaluate the proposed framework.} We conduct extensive experiments based on constructed datasets and state-of-the-art model architectures. The results demonstrate the effectiveness of PFA. 

\end{itemize}

\section{BACKGROUND AND TERMINOLOGY}
In order to clarify the meaning of specific terms used in this paper, and to help readers get an overall understanding of how federated learning works, we briefly describe some important concepts in the Convolutional Neural Network (CNN) and the workflow of FL. 

\begin{figure}[t]
\centering
\includegraphics[width=1.0\columnwidth]{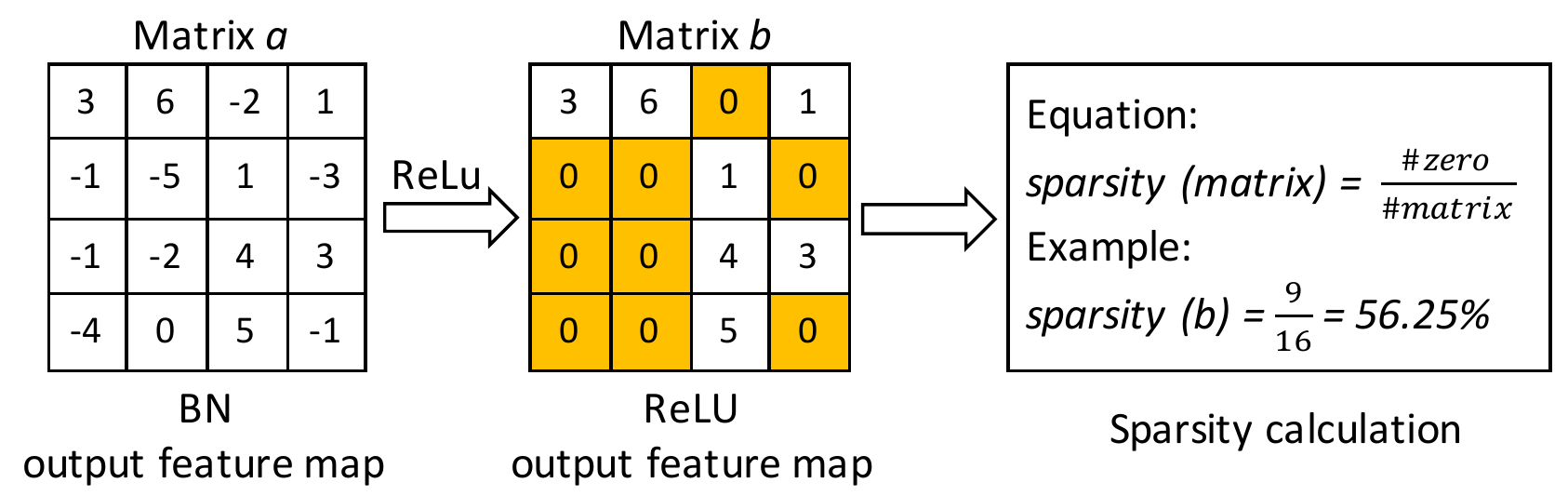} 
\caption{Illustration of the different feature maps and the sparsity calculation.}
\label{fig:sparsity_illustration}
\end{figure}

\subsection{Convolutional Neural Network}
\label{sec:cnn_notation}
By default, our work is based on Convolutional Neural Networks (CNNs) \cite{lecun1998gradient}, which have been widely used in the field of computer vision, such as image classification \cite{simonyan2014very,he2016deep,huang2017densely}, object detection \cite{girshick2014rich,lin2017focal} and media interpretation \cite{vukotic2016multimodal,forgione2018implementation}. A typical CNN may include convolutional (Conv) layers, pooling layers, batch normalization (BN) layers, ReLU activation functions and fully connected (FC) layers, each of which represents a unique type of mathematical operations. By feeding an image into a CNN, the output of each layer can be regarded as a type of \textit{feautre map} with a shape of H$\times$W$\times$C, in which H, W, C represent the height, weight, and the number of channels in the feature map. As illustrated in Figure \ref{fig:sparsity_illustration}, after the processing of the BN and ReLU layer, we can generate the corresponding \textit{BN output feature map} and \textit{ReLU output feature map} (Here we only demonstrate one channel). 

Notice that the ReLU layer can turn all negative inputs into zeros, thus making its output highly sparse. Given a ReLU output feature map, for a certain channel, the \textit{sparsity} is defined as the ratio for the number of zero elements in a matrix. For the example in Figure \ref{fig:sparsity_illustration}, there are 9 zero elements out of 16 elements in the matrix, and thus the \textit{sparsity} is 56.25\%. In this way, we can calculate the sparsity value of each channel and the whole feature map will correspond to a sparsity vector. In the following sections, these terms (layer, channel, sparsity \textit{etc}.) will be used to explain our approach.

\subsection{Federated Learning}

The typical pipeline of FL includes two steps. First, each client trains a model locally with a few epochs and only uploads the parameters/gradients of the model to a central server. Second, the server end coordinates the uploaded information through different aggregation algorithms (e.g., Federated Averaging (FedAvg)), in order to generate a global model that owns the knowledge of each client. The generated global model is finally distributed to each client and may be further trained to iterate the above steps until convergence. During this process, the raw data in each client is not exchanged, thus without leaking the user privacy.

\begin{figure*}[t]
\centering
\includegraphics[width=2.1\columnwidth]{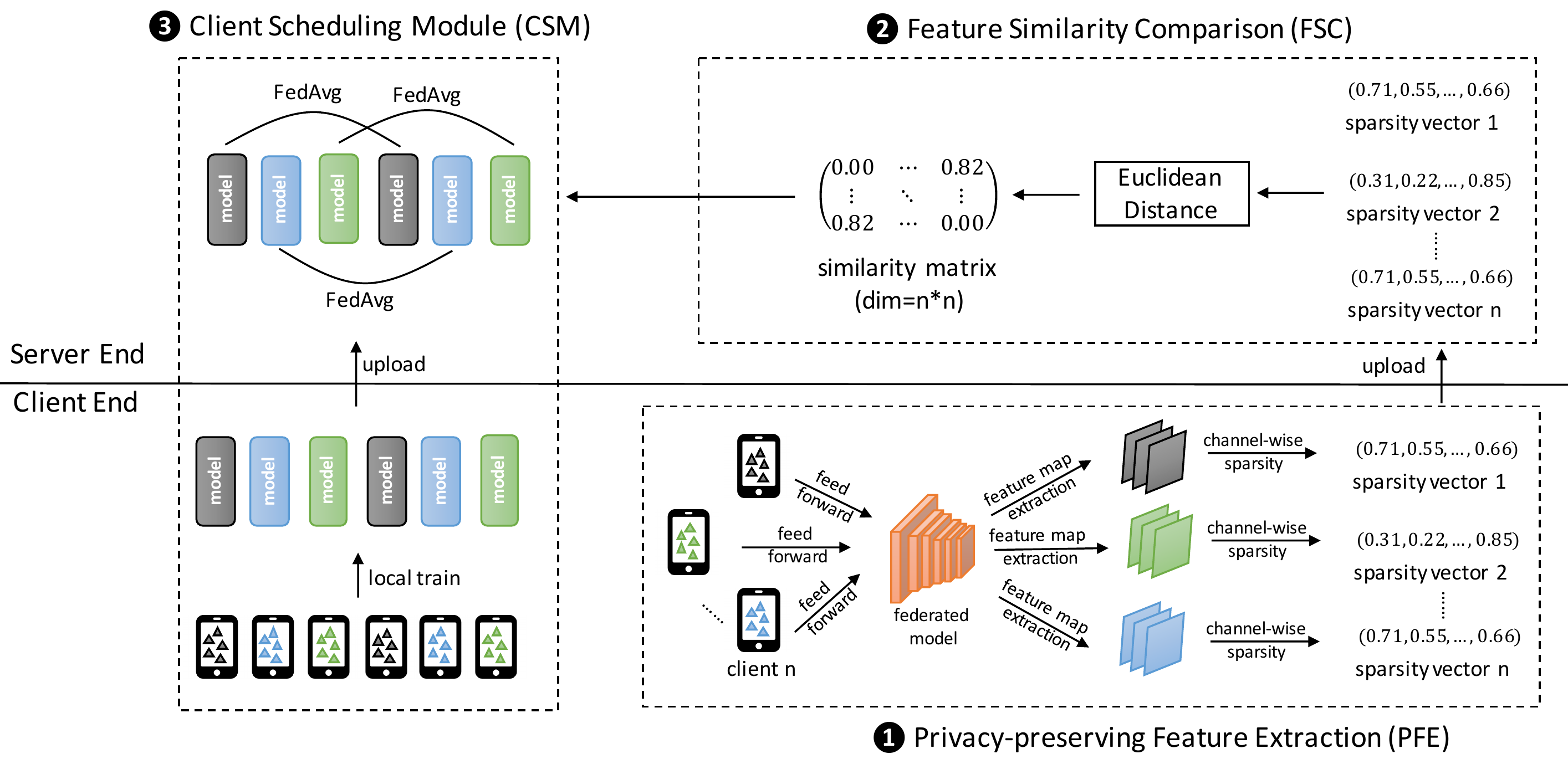} 
\caption{The overview of the proposed framework PFA. Three modules (i.e., PFE, FSC, and CSM) are introduced and sequentially executed to achieve federated adaptation. Concretely, PFE extracts the privacy-preserving representation; FSC uses the representation to generate a similarity matrix between clients and CSM utilizes the matrix to schedule clients by partitioning their corresponding trained models into different groups and conducting group-wise adaptation with the FedAvg algorithm. Note that we do not exchange raw data of clients during the whole process, offering a good privacy protection.}
\label{fig:overview}
\end{figure*}

\section{Goal and challenge}

\begin{table}[]
\centering
\caption{Notations used in this paper.}
\label{tab:notation}
\begin{tabular}{c|l}
\hline
\textbf{Notation} & \textbf{Explanation} \\ \hline \hline
 $M_f$        &      The federated model       \\
 $M_p$      &       The personalized model (i.e., our goal)      \\
 $M_l$         &    The trained local model        \\
 $R$       &      The privacy-preserving feature representation       \\
 $S$        &     A similarity matrix among clients        \\
 $D$       &      The data in a client       \\
 $E$        &     The element of a specific $R_x$        \\
 $N$       &      The number of data in a certain client       \\
 $sp$        &    The sparsity of a ReLU feature map        \\
 $n$         &    The number of clients        \\
 $q$       &      The number of selected channels       \\ \hline
\end{tabular}
\end{table}

This section formulates the goal and existing challenges of PFA. The important notations that will be commonly used later are summarized in Table \ref{tab:notation}. 

\subsection{Problem Formulation}
Starting with a model that is trained by the traditional FL process, PFA first aims to use it to generate a privacy-preserving feature representation vector $R=(R_1, R_2,...,R_n)$ for clients. Here the $R_i$ denotes the feature representation of the $i_{th}$ client and $n$ is the total number of the clients. The representation vector $R$ is then uploaded to the central server and PFA targets at using it to compare the distribution similarity between clients, generating a similarity matrix $S$. In terms of $S$, PFA attempts to schedule clients by partitioning them into different groups, each of which represents a type of data distribution. As a result, each client can benefit from other clients in the group that own data with a similar distribution. 

Based on these symbols, the final goal of PFA can be defined as follows:

\begin{defn}
\label{defn:problem}
\textbf{(Privacy-preserving federated adaptation)}
Suppose $M_{f}$ and $M_{p}^j$ denote the federated global model and the desirable personalized model of the $j_{th}$ client, respectively. The goal of PFA is to group a series of clients $(j_1,j_2,...,j_m)$ that own similar data distribution to $j$, with the help of $M_{f}$, $R$, and $S$ rather than the raw data, in order to adapt $M_{f}$ in a federated manner and generate $M_{p}^j$.
\end{defn}

\subsection{Challenges}

In this work, we aim at exploring a novel federated adaptation approach to achieve personalization. It is nevertheless non-trivial to accomplish this goal. There are at least three challenges that need to be effectively addressed. We list them as follows.
\begin{enumerate}
    \item \textbf{How to ensure user privacy in federated adaptation?} 
    As is known to all, privacy protection is the key concern in FL. Similarly, the adaptation process should also pay much attention to the privacy issue. Specifically, the raw data in clients cannot be transferred or inferred during the whole process.
    
    \item \textbf{How to efficiently distinguish clients with different data distributions?}
    An important point in federated adaptation is that the data of federated clients should come from an identical or similar distribution. Therefore, a similarity metric for client distribution is needed to be developed. Besides, there might be a huge number of clients, which poses a higher demand for an efficient method to accomplish the goal.
    
    \item \textbf{How to accurately and effectively identify and schedule the federated clients?}
    Unlike the local adaptation in which only one client is needed, for federated adaptation, it is crucial to determine which clients should be aggregated and how they can contribute to each other.
    
\end{enumerate}

\section{Our framework: PFA}
We design and develop PFA, a framework to achieve personalization in a federated manner. Figure \ref{fig:overview} depicts the essential modules of the workflow in PFA. Specifically, we introduce three modules to respectively address the aforementioned three challenges:

\begin{itemize}
    \item \textbf{PFE:} a \textbf{P}rivacy-preserving \textbf{F}eature \textbf{E}xtraction module, which takes advantage of the feature map sparsity to represent the distribution property of clients. This module only uses the federated model to conduct a feed forward process, without introducing much computation cost or complex optimizations. The generated sparsity vectors are then uploaded to the server end.
    
    \item \textbf{FSC:} a \textbf{F}eature \textbf{S}imilarity \textbf{C}omparison module, which employs Euclidean Distance to measure the similarity between the extracted sparsity vectors, in order to generate a similarity matrix that can denote the clients' similarity degree. Besides, we further design a one-client-related similarity vector to form the matrix, so as to achieve efficiency.
    
    \item \textbf{CSM:} a \textbf{C}lient \textbf{S}cheduling \textbf{M}odule, which partitions the clients into different groups in terms of the similarity matrix generated by \textbf{FSC} and implements adaptation in a group-wise manner. Concretely, each client needs to upload its corresponding local trained model and the server selectively groups these models and aggregates them by the FedAvg algorithm. 
    
\end{itemize}

Besides, we would like to highlight that different modules are conducted in different ends: \textbf{PFE} is in the client end; \textbf{FSC} is in the server end, and \textbf{CSM} requires both of the ends. In the remainder of the section, we describe in detail our approach for implementing each module. 

\subsection{Privacy-preserving Feature Extraction}
\label{sec:PFE}
The PFE module serves as the first step in the proposed framework. Its objective is to extract a representation for each client that  can not only reflect the distribution property but also protect user privacy. A natural idea is to use the feature map extracted from each client as the representation since it is directly associated with the raw data. However, the information of the feature map is easy to be inverted to generate the original raw data \cite{mahendran2015understanding,fredrikson2015model}, which violates the user privacy (details in \textbf{Section \ref{sec:inversion_analysis}}).

In this work,  we make an important observation: \textit{by feeding the data of a client into a federated model, its intermediate channel sparsity can express data distribution information of the client.} The insight behind it is that usually the sparsity pattern for specific inputs is unique, which suggests that we can use it as the client representation in order to distinguish different data distributions. Here sparsity refers to the ratio of the number of zero elements in a matrix (details in \textbf{Section \ref{sec:cnn_notation}}), which has been widely used to accelerate the inference process of Convolutional Neural Networks (CNNs) \cite{ren2018sbnet,cao2019seernet}. Using sparsity as the representation has the following two advantages: (1) Sparsity can be seen as a type of statistical information, which is intuitively privacy-preserving as it converts the sensitive features of raw data into the simple ratio, making it hard to be inverted (details in \textbf{Section \ref{sec:inversion_analysis}}); (2) Sparsity is smaller than its corresponding feature map, which significantly reduces the communication costs (i.e., uploading them to the server end).

Based on this observation, we attempt to extract the channel-wise sparsity vector as the distribution representation. Specifically, for the $i_{th}$ client, we denote its data as $D_i=(x_1^i,x_2^i,...,x_N^i)$ and conduct a feed forward process to extract the feature maps. Here $N$ represents the total number of the data in the client. Let $F(x_p^i) \in \mathbb{R}^{H \times W \times C}$ be the feature map extracted from a ReLU layer of the federated model given input $x_p^i\in D_i$. For a channel in the layer, we compute the sparsity by the following equation
\begin{equation}
    \label{equation:activation_extract}
    sp(x_p^i)=\frac{\# zero \ elements }{H\times W} 
\end{equation} 
where $sp$ denotes the sparsity. With this equation, we calculate the sparsity of each sample in $D_i$ and average them as the client sparsity for the this channel
\begin{align}
\label{equation:descriptor}
    sp(D_i)=\frac{1}{N} \sum_{k=1}^{N} sp(x_p^i)
\end{align}
In this way, we randomly select $q$ channels $(c_1, c_2, ..., c_q)$ from the federated model and extract their corresponding client sparsity to form a sparsity vector, which is considered as the privacy-preserving representation $R_i$ for the $i_{th}$ client.
Based on these steps, each client can generate a representation and upload it to the server end for later similarity comparison.

\subsection{Feature Similarity Comparison}
\label{sec:FSC}

The FSC module compares the representations (i.e., sparsity vectors) extracted through PFE by computing the distance among them. Here we adopt Euclidean Distance \cite{danielsson1980euclidean} as the metric due to its simplicity and prevalence. 

Concretely, for the representation $R_i$ of the $i_{th}$ client and representation $R_j$ of the $j_{th}$ client, their similarity can be measured by 
\begin{align}
\label{equation:similarity}
    sim(R_i,R_j)=\sqrt{\sum_{k=1}^{q} (E_i^k-E_j^k)^2}
\end{align}
where $E_i^k$ and $E_j^k$ denote the $k_{th}$ element in  $R_i$ and $R_j$, respectively. Each client representation has $q$ elements since we pick out $q$ channels in above steps. Based on Eq. \ref{equation:similarity}, we calculate the similarity of any of two clients and generate a similarity matrix $S$, where $S_{ij}$ represents the distribution similarity degree between the $i_{th}$ and the $j_{th}$ client.

Although the above method is effective in comparing the similarity, it may introduce unacceptable computation budgets when the number of clients is huge. For example, we need to calculate $C_{100,000}^2$ times Euclidean Distance if there are 100,000 clients involved in the adaptation phase, which is inefficient and will largely slow the overall adaptation speed.

To overcome the efficiency challenge for fast comparison, we further propose to only calculate the similarity with respect to one client rather than the whole clients. Specifically, we first randomly pick out a representation $R_z$ of the $z_{th}$ client, and then compute the Euclidean Distance between it and other representations, generating a final similarity vector 
\begin{align}
\label{equation:}
   \big( sim(R_z,R_1),sim(R_z,R_2),...,sim(R_z,R_n) \big)
\end{align}
We observe that in fact, this vector is enough to judge the similarity of clients. On one hand, if the value of $sim(R_z,R_g)$ is low, we believe the $g_{th}$ client has similar data distribution to the $z_{th}$ client. On the other hand, if the value of $sim(R_z,R_t)$ and $sim(R_z,R_u)$ is close, the two corresponding clients (the $t_{th}$ and $u_{th}$) can be considered as sharing the similar data distribution. This judgement is reasonable because generally the option of data distributions is limited (e.g., the number of classes or the background is not infinite in our vision scenario). With the two principles, we can obtain the similarity of any two clients and form the final similarity matrix $S$. In this way, given 100,000 clients, the total computation budgets are reduced from $C_{100,000}^2$ to $99,999$ times Euclidean Distance calculation, significantly accelerating the comparison process. Experiments in Section \ref{sec:extract_strategy} confirm the effectiveness of the proposed efficient scheme.

\begin{table}[]
\centering
\caption{Statistics of our simulated datasets.}
\label{tab:simulated_dataset}
\begin{tabular}{ll|rrr}
\hline
\multicolumn{2}{c}{\multirow{2}{*}{\textbf{Dataset}}}                  & \multicolumn{1}{c}{\textbf{Client}} & \multicolumn{1}{c}{\textbf{\#training}} & \multicolumn{1}{c}{\textbf{\#testing}} \\
\multicolumn{2}{c}{}                                                        & \multicolumn{1}{c}{\textbf{number}}          & \multicolumn{1}{c}{\textbf{sample}}              & \multicolumn{1}{c}{\textbf{sample}}             \\ \hline
\multicolumn{1}{l|}{\multirow{5}{*}{\textit{Cifar10}}}     & \textit{type1} & 1,2,3,4,5                           & 5*100                                   & 5*100                                  \\
\multicolumn{1}{l|}{}                                      & \textit{type2} & 6,7,8,9,10                          & 5*100                                   & 5*100                                  \\
\multicolumn{1}{l|}{}                                      & \textit{type3}          & 11,12,13,14,15                      & 5*100                                   & 5*100                                  \\
\multicolumn{1}{l|}{}                                      & \textit{type4}          & 16,17,18,19,20                      & 5*100                                   & 5*100                                  \\
\multicolumn{1}{l|}{}                                      & \textit{type5}          & 21,22,23,24,25                      & 5*100                                   & 5*100                                  \\ \hline
\multicolumn{1}{l|}{\multirow{4}{*}{\textit{Office-Home}}} & \textit{Ar}    & 1,2,3,4,5                           & 5*291                                   & 5*97                                   \\
\multicolumn{1}{l|}{}                                      & \textit{Cl}    & 6,7,8,9,10                          & 5*523                                   & 5*174                                  \\
\multicolumn{1}{l|}{}                                      & \textit{Pr}    & 11,12,13,14,15                      & 5*532                                   & 5*177                                  \\
\multicolumn{1}{l|}{}                                      & \textit{Rw}    & 16,17,18,19,20                      & 5*522                                   & 5*174                                  \\ \hline
\end{tabular}
\end{table}

\subsection{Client Scheduling Module}
\label{sec:CSM}

The client scheduling module aims at picking out  a series of clients with similar data distribution and utilizing them to cooperatively personalize the federated model. Thanks to the similarity matrix generated by our FSC module, we are able to easily compare the distribution property of different clients. In practice, there are a large number of clients in the FL process, which indicates that every client might find at least one client with the desirable distribution. Inspired by this, the scheduling module attempts to partition the whole clients into different groups, each of which represents a certain data distribution. As a result, the adaptation process would be operated in each group, getting rid of the influence of other unrelated clients.

Specifically, the pipeline includes the following three steps:
\begin{enumerate}
    \item In the client end, each device should first train the federated model for a few epochs, in order to incorporate the client-specific knowledge into the model. The training objective is formulated as follows
    
    \begin{equation}
    M_l^i=\arg \min_{M_{f}} \mathcal{L}(D_i;M_{f})
    \end{equation}
    where $\mathcal{L}$ represents the loss calculated by the loss function (e.g., cross-entropy loss) used to optimize parameters. $M_l^i$ is the trained local model for the $i_{th}$ client.
    
    \item The local trained models $M_{l}^1, M_l^2,...,M_l^n$ are then uploaded to the server end and grouped according to the similarity matrix (i.e., if the distribution of two clients is similar, their corresponding local models should be clustered).
    
    \item We finally conduct the FedAvg algorithm \cite{mcmahan2017communication}, which has been proved effective in aggregating the model knowledge especially for the IID distribution, to accomplish the federated adaptation process. Here in each group, the data distribution can be seen as IID and thus it is desirable to use this algorithm.
    As shown in Figure \ref{fig:overview}, models with similar distributions (denoted by colors) collaborate with each other using the FedAvg algorithm, making the resulting model customized to a certain distribution. 
\end{enumerate}
The three steps may iterate several times until the aggregated model in each group well fits the corresponding distribution. In this way, each group will finally share an adapted model and this local federated model can be considered as the personalization result. We believe the resulting model is better than the local adapted model because more useful knowledge is aggregated, mitigating the overfitting problem and bias existed in the local adaptation pipeline.

\section{EVALUATION}
Our evaluation is driven by the following research questions (RQs).

\textbf{RQ1:} What's the performance of PFA? How does it compare to other personalization methods?

\textbf{RQ2:}  How effective is the sparsity-based representation in distinguishing clients with different data distributions?  

\textbf{RQ3:} Can the extracted sparsity vectors defend the inversion attack and ensure user privacy?

\textbf{RQ4:}  How do different extraction strategies of the privacy-preserving representation affect the distribution similarity comparison?

\begin{figure}[t]
\centering
\includegraphics[width=1.0\columnwidth]{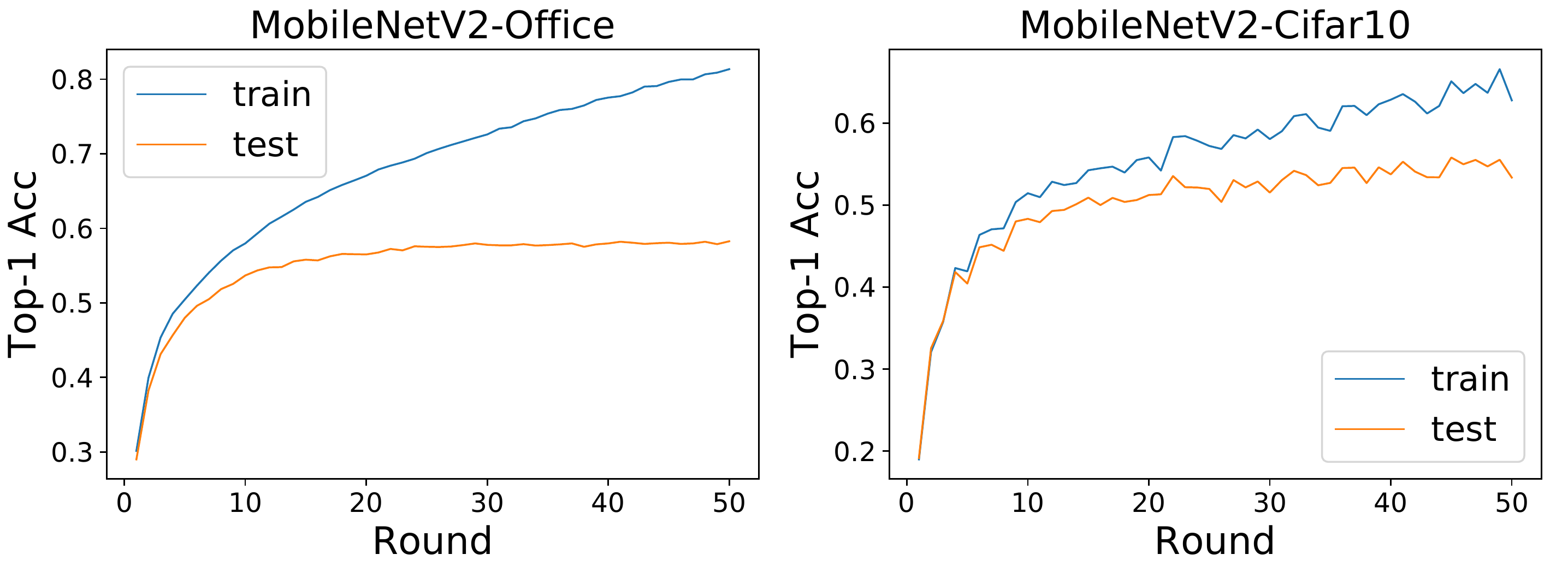} 
\caption{The training process for the two types of datasets on MobileNetV2.}
\label{fig:training_process}
\end{figure}

\begin{table*}[]
\centering
\caption{Office-Home results for the baseline and different personalization methods on VGG-11.}
\label{tab:result_vgg}
\begin{tabular}{c|ccccc|ccccc}
\hline
\multirow{2}{*}{\textbf{Method}} & \multicolumn{5}{c|}{\textbf{Ar}}                                                   & \multicolumn{5}{c}{\textbf{Cl}}                                                                              \\ \cline{2-11} 
                                 & client1        & client2        & client3        & client4        & client5        & client6        & client7        & client8        & \multicolumn{1}{l}{client9} & \multicolumn{1}{l}{client10} \\ \hline
Baseline                         & 40.21          & 50.52          & 59.79          & 51.55          & 53.06          & 42.53          & 48.85          & 48.85          & 48.85                       & 41.81                        \\
Fine-tune                        & 43.30          & 56.70          & 61.86          & 54.64          & 55.10          & 52.30          & 53.45          & 56.32          & 51.15                       & 49.72                        \\
KD                               & 43.30          & 59.79          & 63.92          & 58.76          & 55.10          & 50.00          & 55.75          & 59.20          & 54.60                       & 53.11                        \\
EWC                              & 43.30          & 54.64          & 62.89          & 58.76          & 55.10          & 51.72          & 54.02          & 59.20          & 53.45                       & 52.54                        \\
Ours                             & \textbf{43.30} & \textbf{61.86} & \textbf{63.92} & \textbf{60.82} & \textbf{60.20} & \textbf{59.20} & \textbf{59.20} & \textbf{63.22} & \textbf{59.20}              & \textbf{53.11}               \\ \hline
\multirow{2}{*}{\textbf{Method}} & \multicolumn{5}{c|}{\textbf{Pr}}                                                            & \multicolumn{5}{c}{\textbf{Rw}}                                                                              \\ \cline{2-11} 
                                 & client11       & client12       & client13       & client14       & client15       & client16       & client17       & client18       & client19                    & client20                     \\ \hline
Baseline                         & 73.45          & 71.75          & 68.93          & 76.84          & 73.89          & 61.49          & 66.09          & 61.49          & 63.22                       & 68.18                        \\
Fine-tune                        & 79.10          & 77.40          & 75.71          & 77.40          & 77.22          & 63.22          & 68.97          & 65.52          & 63.22                       & 68.75                        \\
KD                               & 79.10          & 77.40          & \textbf{78.53} & 78.53          & 78.33          & 63.22          & 71.84          & 64.94          & 65.52                       & 69.89                        \\
EWC                              & 79.10          & 80.79          & 76.84          & 77.40          & 79.44          & 63.22          & 70.69          & \textbf{65.52} & 63.22                       & 69.32                        \\
Ours                             & \textbf{81.36} & \textbf{80.79} & 76.84          & \textbf{79.10} & \textbf{81.67} & \textbf{65.52} & \textbf{72.99} & 64.94          & \textbf{68.39}              & \textbf{71.02}               \\ \hline
\end{tabular}
\end{table*}

\begin{table*}[]
\centering
\caption{Office-Home results for the baseline and different personalization methods on MobileNetV2.}
\label{tab:result_mobilenet}
\begin{tabular}{c|ccccc|ccccc}
\hline
\multirow{2}{*}{\textbf{Method}} & \multicolumn{5}{c|}{\textbf{Ar}}                                                   & \multicolumn{5}{c}{\textbf{Cl}}                                                                              \\ \cline{2-11} 
                                 & client1        & client2        & client3        & client4        & client5        & client6        & client7        & client8        & \multicolumn{1}{l}{client9} & \multicolumn{1}{l}{client10} \\ \hline
Baseline                         & 42.27          & 43.30          & 49.48          & 44.33          & 51.02          & 51.15          & 51.15          & 51.15          & 47.70                       & 42.37                        \\
Fine-tune                        & 43.30          & 49.48          & 54.64          & 44.33          & 50.00          & 56.32          & 56.32          & 58.05          & 56.32                       & 50.28                        \\
KD                               & 42.27          & 51.55          & 52.58          & 46.39          & 55.10          & 56.32          & 56.32          & 60.34          & 55.75                       & 54.24                        \\
EWC                              & 43.30          & 53.61          & 54.64          & 47.42          & \textbf{55.10} & 55.75          & 55.75          & 60.34          & 58.62                       & 52.54                        \\
Ours                             & \textbf{43.30} & \textbf{54.64} & \textbf{54.64} & \textbf{52.58} & 52.00          & \textbf{62.64} & \textbf{62.64} & \textbf{60.34} & \textbf{63.79}              & \textbf{57.06}               \\ \hline
\multirow{2}{*}{\textbf{Method}} & \multicolumn{5}{c|}{\textbf{Pr}}                                                            & \multicolumn{5}{c}{\textbf{Rw}}                                                                              \\ \cline{2-11} 
                                 & client11       & client12       & client13       & client14       & client15       & client16       & client17       & client18       & client19                    & client20                     \\ \hline
Baseline                         & 71.19          & 70.62          & 64.41          & 69.49          & 67.22          & 59.20          & 61.49          & 60.92          & 62.07                       & 61.36                        \\
Fine-tune                        & 75.14          & 74.01          & 67.23          & 74.01          & 73.33          & 60.34          & 61.49          & 60.92          & 59.77                       & 63.07                        \\
KD                               & 73.45          & 72.88          & 64.41          & 75.14          & 72.78          & 59.20          & 62.64          & 62.07          & 62.07                       & 65.34                        \\
EWC                              & 75.71          & 74.58          & 68.36          & 74.58          & 71.67          & 61.49          & 62.64          & 62.64          & 60.92                       & 66.48                        \\
Ours                             & \textbf{80.79} & \textbf{78.53} & \textbf{74.58} & \textbf{77.97} & \textbf{81.67} & \textbf{61.49} & \textbf{62.64} & \textbf{62.64} & \textbf{62.64}              & \textbf{66.48}               \\ \hline
\end{tabular}
\end{table*}

\subsection{Experimental Setup}
\label{sec:experiemnt_setup}

The experimental setup includes the construction of the practical FL datasets, the used models, the implementation details and the compared methods.

\textbf{Simulated FL datasets.} In real-world applications, the distribution divergence can be categorized into two types. The first one is the \textit{class-imbalance} condition, where the statistical information of the data in each client may be extremely different. Take the image classification task as an example, some clients may mainly hold samples of the ``cat'' class while others may have few ``cat'' images but a large number of ``dog'' images. Similar to previous work \cite{mcmahan2017communication}, we simulate this situation by the public Cifar10 dataset \cite{krizhevsky2009learning}, which contains 10 different image classes. Specifically, we simulate 25 clients and every 5 clients belongs to a type of distribution. Therefore, totally we have 5 types of distribution, each of which owns two disjoint classes of Cifar10. Considering that the client end may own limited data, we only randomly select 100 samples as the training set and testing set for each client. 

In addition to the \textit{class-imbalance} condition,  the \textit{background-difference} scenario is also commonly seen in practice. For example, photos taken from different environments sometimes are hard to be classified by neural networks although the main object is identical, which is the focus in the field of domain adaptation \cite{pan2009survey}. Under this circumstance, we use the Office-Home dataset \cite{venkateswara2017deep} for simulation. Office-Home contains 15,500 images with four domains: Artistic images (Ar), Clipart images (Cl), Product images (Pr) and Real-World images (Rw). Each domain has 65 identical object categories but different background distributions. Concretely, we divide each domain into 5 parts and each client owns one part. For each part, we further partition it into the training data (80\%) and the testing data (20\%) since this dataset has no official train/test partition. According to the setting, totally we have 20 clients with 4 types of data distribution. Because the data in each domain of Office-Home is limited, we do not need to implement sampling as we do for Cifar10. We illustrate the detailed statistics in Table \ref{tab:simulated_dataset}.

\textbf{Models.} We evaluate the proposed approach on two widely used model architectures: VGGNet \cite{simonyan2014very} and MobileNetV2 \cite{sandler2018mobilenetv2}. VGGNet is a vanilla convolutional neural network with direct connections across Conv layers. We modify the architecture by replacing the original fully connected layers with a global average pooling layer in order to fit the data scale in our FL setting. MobileNetV2 is a typical convolutional neural network designed for the mobile end, which is suitable for our scenario. For each architecture, we use the smallest variant (i.e., VGG-11 and MobileNetV2-0.25) considering the resource constraint of clients. Besides, we use their pre-trained versions (i.e., trained with the ImageNet dataset \cite{deng2009imagenet}) before applying to the actual tasks to accelerate convergence.

\textbf{PFA implementation.} 
Since it is hard to conduct experiments in real-world FL scenarios, we simulate and operate our experiments in a server that has 4 GeForce GTX 2080Ti GPUs, 48 Intel Xeon CPUs, and 128GB memory. We implement PFA in Python with PyTorch \cite{paszke2019pytorch} and all the experiments are conducted based on the above datasets and models. 

The concrete parameter settings are as follows: In the federated learning and federated adaptation process, the learning rate is set to 0.01, with a momentum of 0.5. The training is conducted for 50 rounds and 30 rounds for the two processes, respectively. In the feature extraction phase, we randomly pick out 30 channels and extract their sparsity to form the sparsity vector. The extraction location is the first ReLU layer of the federated model. Other extraction examples will be further displayed and analyzed in Section \ref{sec:extract_strategy}. 

\textbf{Compared methods.} 
We compare the FL baseline and other three personalization methods to our PFA. 
\begin{enumerate}
    \item \textit{Baseline.} The global federated trained model can be used to test the performance of all clients, which we consider as the baseline. 
    
    \item \textit{Fine-tuning Adaptation \cite{wang2019federated}.} Fine-tuning technique is a popular paradigm to achieve transfer learning \cite{tajbakhsh2016convolutional}. In the context of FL, this adaptation is used to retrain all parameters of a trained federated model on the participant’s local training data.
    
    \item \textit{KD Adaptation \cite{yu2020salvaging}.} Knowledge Distillation (KD) \cite{hinton2015distilling}
    extracts information from a “teacher” model into a “student” model. Here we treat the federated model as the teacher and the adapted model as the student in order to implement the knowledge distillation process. The distilled model can be regarded as the personalized model.
    
    \item \textit{EWC Adaptation \cite{yu2020salvaging}.} Elastic Weight Consolidation (EWC) \cite{kirkpatrick2017overcoming} is used to overcome the catastrophic forgetting problem \cite{french1999catastrophic}. In our scenario, we aim to utilize it to force the federated model to be adapted for the client data while preserving the original important knowledge, with the purpose of mitigating the overfitting problem to some extent.
\end{enumerate}
Note that all these methods are conducted in a single device, failing to borrow the useful knowledge existed in other devices.

\begin{table}[]
\centering
\caption{Achieved accuracy (\%) of different methods on VGG-11. \textit{Here Ar, Cl, Pr, Rw represent different domains.}}
\label{tab:domain_vgg}
\begin{tabular}{@{}cccccc@{}}
\toprule
\textbf{Method} & \textbf{Ar}             & \textbf{Cl}             & \textbf{Pr}             & \textbf{Rw}             & \textbf{Avg}   \\ \midrule
Baseline        & 51.03          & 46.18          & 72.97          & 64.09          & 58.57          \\
Fine-tune       & 54.32          & 52.59          & 77.37          & 65.94          & 62.55          \\
KD              & 56.17          & 54.53          & 78.38          & 67.08          & 64.04          \\
EWC             & 54.94          & 54.19          & 78.71          & 66.39          & 63.56          \\
Ours            & \textbf{58.02} & \textbf{58.79} & \textbf{79.95} & \textbf{68.57} & \textbf{66.33} \\ \bottomrule
\end{tabular}
\end{table}

\begin{table}[]
\centering
\caption{Achieved accuracy (\%) of different methods on MobileNetV2.}
\label{tab:domain_mobilenet}
\begin{tabular}{@{}cccccc@{}}
\toprule
\textbf{Method} & \textbf{Ar}             & \textbf{Cl}             & \textbf{Pr}             & \textbf{Rw}             & \textbf{Avg}   \\ \midrule
Baseline        & 46.08          & 48.70          & 68.59          & 61.01          & 56.09          \\
Fine-tune       & 48.35          & 55.00          & 72.74          & 61.12          & 59.30          \\
KD              & 49.58          & 56.59          & 71.73          & 62.26          & 60.04          \\
EWC             & 50.81          & 56.94          & 72.98          & 62.83          & 60.89          \\
Ours            & \textbf{51.43} & \textbf{61.18} & \textbf{78.71} & \textbf{63.18} & \textbf{63.62} \\ \bottomrule
\end{tabular}
\end{table}

\subsection{RQ1: Overall Results}
Before applying personalization, we first need to implement traditional FL to generate a federated model. Towards our simulated FL datasets, we observe that using FL alone cannot achieve a good performance. To give a more direct understanding, we visualize the training process for the two types of datasets on MobileNetV2. As shown in Figure \ref{fig:training_process}, for Office-Home, the gap between the training and testing accuracy is extremely large, suggesting that the federated model cannot be generalized well. For Cifar10, we can clearly see the instability both in the training and testing phase, which indicates that the model is hard to converge during the FL process. These two phenomena further demonstrate the necessity to personalize the federated model for better performance.

\textbf{Personalization on Office-Home.}
Given a federated model, we test its accuracy as the baseline and conduct personalization using 4 methods (i.e., fine-tuning, KD, EWC, and PFA). The detailed results of each client are shown in Table \ref{tab:result_vgg} and Table \ref{tab:result_mobilenet}. For most clients, the personalization performance (i.e., accuracy) achieved by PFA outperforms other methods by a large margin, both on the VGG-11 and MobileNetV2-0.25 model. Specifically, For VGG-11, PFA demonstrates its superiority on 18 out of 20 clients with up to 6.90\% accuracy improvement (for \textit{client6}). For MobileNetV2-0.25, 19 out of 20 clients gain more benefit from the proposed approach compared to the baseline and other personalization methods. Notably, for \textit{client15}, PFA exceeds other methods by 8.34\%, bringing a significantly positive effect to the user experience. 

In addition, we average the accuracy of each domain and report the domain performance in Table \ref{tab:domain_vgg} and Table \ref{tab:domain_mobilenet}. We summarize the following conclusions: (1) The benefits of personalization vary from different domains. For example, personalization contributes less on clients of the \textit{Rw} domain. We believe this is because the \textit{Rw} domain includes more images with general features, which can also be facilitated by other domain data during the FL process although their distributions are different. (2) For the same domain, the model architecture can largely affect the personalization performance of PFA. For instance, the improvement using PFA on VGG-11 is far less than the MobilenetV2-0.25 under the \textit{Pr} domain. Overall, the average improvement is over 2\% on both models, which confirms the effectiveness of the proposed approach.

\begin{figure}[t]
\centering
\includegraphics[width=1.0\columnwidth]{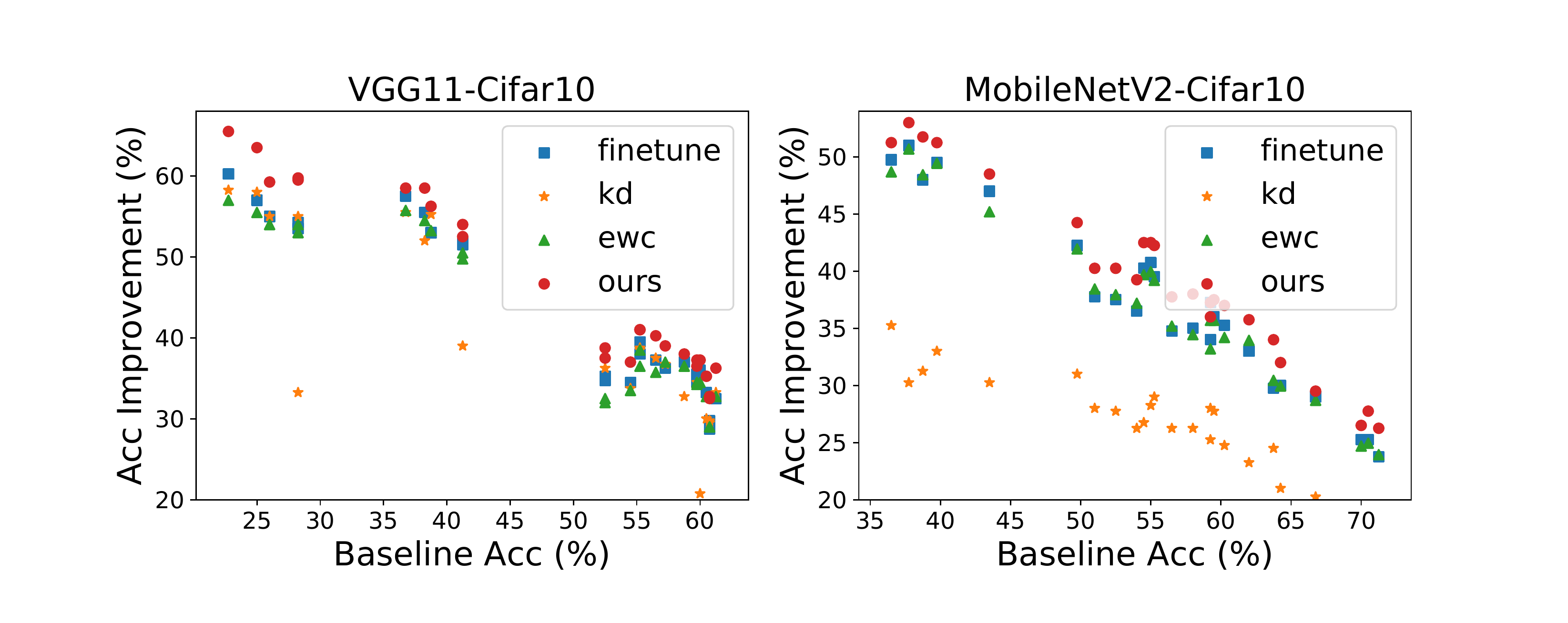} 
\caption{Accuracy improvement achieved by different methods on Cifar10.}
\label{fig:result_cifar}
\end{figure}

\textbf{Personalization on Cifar10.}
Figure \ref{fig:result_cifar} illustrates the results on Cifar10. Here we use the accuracy improvement to the baseline as the measurement of the performance. Each point in the figure represents a client model generated by a certain personalization method. As shown in the figure, all the personalization methods can significantly outperform the baseline performance, suggesting that the federated model on the \textit{class-imbalance} setting has greater demand to be personalized. Among these methods, our PFA can achieve consistently better accuracy, especially when the baseline accuracy is low. Besides, it is noteworthy that the KD method can not perform well. The reason may be that this method uses the federated model as a ``teacher'' while the ``teacher'' itself is not good enough. In summary, PFA shows its superiority due to federated adaptation.

\begin{table}[]
\centering
\caption{The number of parameters for uploading.}
\label{tab:param_upload}
\begin{tabular}{@{}cccccc@{}}
\toprule
\textbf{Item}    & VGG11                     & MobileNetV2               & 10\_sp            & 30\_sp             & 50\_sp             \\ \midrule
\textbf{\#Params} & \multicolumn{1}{c}{9.23M} & \multicolumn{1}{c}{0.13M} & \multicolumn{1}{c}{40B} & \multicolumn{1}{c}{120B} & \multicolumn{1}{c}{200B} \\ \bottomrule
\end{tabular}
\end{table}

\textbf{Communication costs.}
Our approach requires uploading some sparsity vectors to the server, which would introduce other communication costs. Table \ref{tab:param_upload} details the number of parameters for uploading. Here the ``xx\_sp'' denotes the size of the sparsity vector. Obviously, compared to the model itself, the parameters of the sparsity information are negligible, indicating that the overall communication costs would not increase much.

\begin{table}[]
\centering
\caption{Achieved accuracy (\%) on different policies of client selection.}
\label{tab:select_policy}
\begin{tabular}{@{}cccc@{}}
\toprule
\multirow{2}{*}{\textbf{\begin{tabular}[c]{@{}c@{}}Client\\ number\end{tabular}}} & \multirow{2}{*}{\textbf{\begin{tabular}[c]{@{}c@{}}Federated\\ learning\end{tabular}}} & \multirow{2}{*}{\textbf{\begin{tabular}[c]{@{}c@{}}Random\\ selection\end{tabular}}} & \multirow{2}{*}{\textbf{\begin{tabular}[c]{@{}c@{}}Sparsity-based\\ selection\end{tabular}}} \\
                                                                                  &                                                                                        &                                                                                                 &                                                                                                         \\ \midrule
client1                                                                           & 42.27                                                                                  & 39.18                                                                                           & \textbf{43.30}                                                                                                   \\
client2                                                                           & 43.30                                                                                  & 39.18                                                                                           & \textbf{54.64}                                                                                                   \\
client3                                                                           & 49.48                                                                                  & 48.45                                                                                           & \textbf{54.64}                                                                                                  \\
client4                                                                           & 44.33                                                                                  & 47.42                                                                                           &\textbf{52.58}                                                                                                   \\
client5                                                                           & 51.02                                                                                  & 52.00                                                                                          & \textbf{52.00}                                                                                                   \\
client6                                                                           & 51.15                                                                                  & 52.30                                                                                           & \textbf{62.64}                                                                                                   \\
client7                                                                           & 51.15                                                                                  & 51.72                                                                                           & \textbf{62.07}                                                                                                   \\
client8                                                                           & 51.15                                                                                  & 55.75                                                                                           & \textbf{60.34}                                                                                                   \\
client9                                                                           & 47.70                                                                                  & 52.87                                                                                           & \textbf{63.79}                                                                                                   \\
client10                                                                          & 42.37                                                                                  & 50.28                                                                                           & \textbf{57.06}                                                                                                   \\
client11                                                                          & 71.19                                                                                  & 66.67                                                                                           & \textbf{80.79}                                                                                                   \\
client12                                                                          & 70.62                                                                                  & 66.67                                                                                           & \textbf{78.53}                                                                                                   \\
client13                                                                          & 64.41                                                                                  & 63.28                                                                                           & \textbf{74.58}                                                                                                   \\
client14                                                                          & 69.49                                                                                  & 66.10                                                                                           & \textbf{77.97}                                                                                                   \\
client15                                                                          & 67.22                                                                                  & 68.89                                                                                           & \textbf{81.67}                                                                                                   \\
client16                                                                          & 59.20                                                                                  & 58.05                                                                                           & \textbf{61.49}                                                                                                   \\
client17                                                                          & 61.49                                                                                  & 56.32                                                                                           & \textbf{62.64}                                                                                                   \\
client18                                                                          & 60.92                                                                                  & 57.47                                                                                           & \textbf{62.64}                                                                                                   \\
client19                                                                          & 62.07                                                                                  & 60.34                                                                                           & \textbf{62.64}                                                                                                   \\
client20                                                                          & 61.36                                                                                  & 57.39                                                                                           & 6\textbf{6.48}                                                                                                   \\ \midrule
Avg                                                                               & 56.09                                                                                  & 55.52                                                                                           & \textbf{63.62}                                                                                                   \\ \bottomrule
\end{tabular}
\end{table}

\subsection{RQ2: Effectiveness of the Sparsity-based Representation}
\label{sec:effectivenss_sparsity}
The sparsity-based representation is a key component that can be used to distinguish clients with different distributions. Thus it is important to assess its effectiveness. Specifically, we evaluate this representation by comparing the following three policies.
\begin{itemize}
    \item \textit{Federated learning.} We conduct the traditional FL process as the baseline. This also can be considered as selecting the whole clients to implement federated adaptation.
    
    \item \textit{Random client selection for federated adaptation.} We randomly pick out clients to conduct federated adaptation, in order to observe the mutual influence among randomly selected clients. 
    
    \item \textit{Sparsity-based client selection for federated adaptation.} We select clients based on the uploaded sparsity representation. Clients with a high similarity of the representation are aggregated to accomplish federated adaptation.
\end{itemize}

Here we use the MobileNetV2 on Office-Home as an example and each policy is implemented based on the setting. The detailed accuracy results are summarized in Table \ref{tab:select_policy}. From the table, we can clearly see that our sparsity-based selection policy performs best for all the 20 clients by a large margin. On average, the accuracy improvement is 7.52\% and 8.10\% for federated learning and random selection, respectively, which validates the effectiveness of the extracted sparsity vectors. Besides, we notice that the random selection cannot perform well, even worse than the traditional federated learning on the average performance. This further demonstrates that the federated clients should be selective in the context of the adaptation scenario.

\begin{figure}[t]
\centering
\includegraphics[width=1.0\columnwidth]{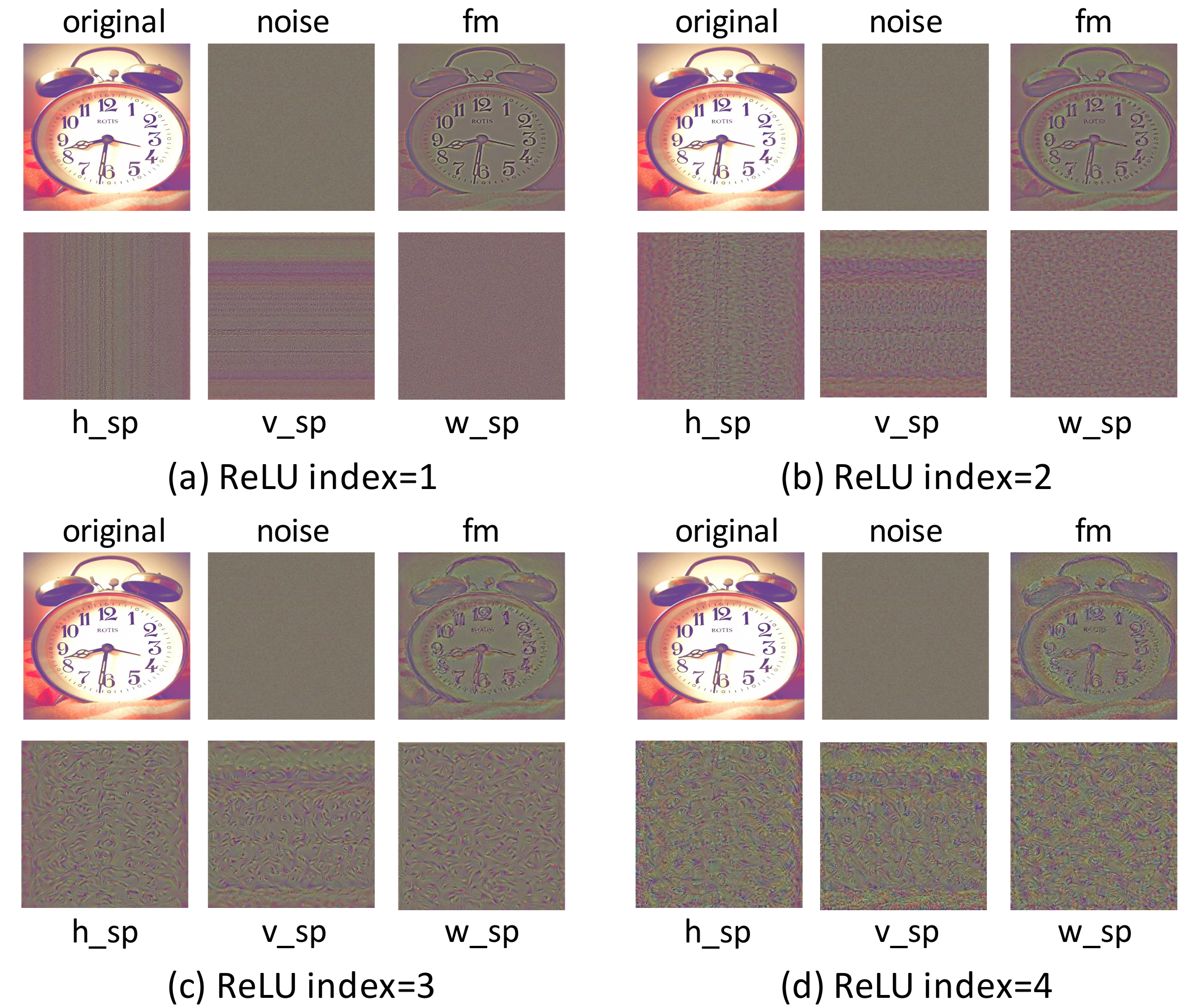} 
\caption{Inversion attack on different properties.}
\label{fig:inversion_attack}
\end{figure}

\subsection{RQ3: Privacy Analysis}
\label{sec:inversion_analysis}
In addition to confirming the effectiveness of the proposed sparsity-based representation, whether this representation can protect user privacy should also be considered. This part provides an analysis with respect to privacy. Concretely, we explore if the sparsity vector can be inverted to generate the original image using the existing attack strategy \cite{fredrikson2015model}. 

We illustrate the results in Figure \ref{fig:inversion_attack}. Four properties (i.e., \textit{fm}, \textit{h\_sp}, \textit{v\_sp}, \textit{w\_sp}) are utilized to implement the inversion attack at different locations (i.e., ReLU index). Here \textit{fm} represents the feature map. \textit{h\_sp}, \textit{v\_sp}, \textit{w\_sp} respectively denote the horizontal-level, vertical-level and whole sparsity of the feature map. Note that we use the sum of the sparsity as the property since a single sparsity value is non-differentiable. From the figure, we can observe that: (1) The \textit{fm} can be easily attacked by inversion since the generated image has abundant features of the original image (e.g., the concrete time in the clock); (2) Only some lines are visualized by inverting the \textit{h\_sp} and \textit{v\_sp}, which cannot reflect any useful information; (3) With the \textit{w\_sp}, attackers fail to inspect any features and the inverted image is just like noises, significantly ensuring the user privacy; (4) The deeper the layer is, the harder we can invert the properties. For example, when it comes to the $4_{th}$ ReLU index, the generated image from all of the properties tends to be noises. However, even if for the first ReLU layer, it is impossible to invert useful knowledge from the sparsity-based representation, which validates its ability to guarantee the confidentiality of user data.

\begin{figure*}[t]
\centering
\includegraphics[width=2.1\columnwidth]{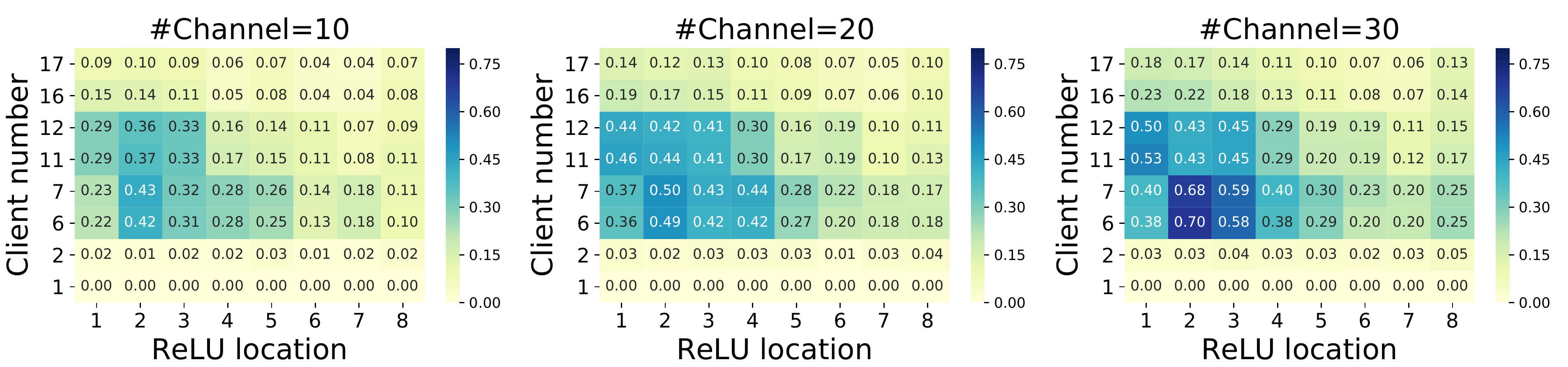} 
\caption{Visualization of the influence on different extraction strategies. }
\label{fig:extract_strategy}
\end{figure*}

\subsection{RQ4: Influence on Different Extraction Strategies}
\label{sec:extract_strategy}
In aforementioned implementation details, we pick out 30 channels' sparsity from the first ReLU layer as the extracted representation. However, there are other options for the extraction. Therefore, we make an in-depth analysis of different extraction strategies, so as to discover the best solution to facilitate personalization. 

Specifically, we first select 8 clients (i.e., 1,2,6,7,11,12,16,17) from the simulated Office-Home dataset and every 2 clients comes from the same distribution. Then for each client, we extract the sparsity vector from each ReLU layer of the federated VGG-11 model. The size (i.e., number of channels) of the vector varies from 10, 20, and 30. Finally, we calculate the Euclidean Distance among these extracted vectors as the metric of client similarity. Here we use \textit{client1} as the base and compare the similarity between it and other clients (i.e., the efficient scheme stated in \textbf{Section \ref{sec:FSC}}). As shown in Figure \ref{fig:extract_strategy}, the heat map is utilized to visualize the similarity degree, where the darker color represents the higher divergence. We find two interesting phenomena. First, the representation from shallow layers performs better than deep layers. This is reasonable because the sparsity of deep layers is extremely high (sometimes can exceed 90\% as illustrated in \cite{cao2019seernet}), making it hard to express the input patterns. Another observation is that more channels contribute to distinguishing clients with different data distributions and in this experiment 30 channels are enough to achieve a good performance.

In summary, the effectiveness of the sparsity depends on the extraction location and quantity. We confirm by experiments that extracting more channels from shallower layers would significantly benefit the final distinguishment.

\section{Discussions}
This section summarizes some limitations of PFA and discusses possible future directions.

\textbf{More model architectures.}
Currently, our framework only targets the CNN-based architecture that is widely used in computer vision tasks. Considering there are many other architectures such as recurrent neural networks (RNNs) and graph convolutional networks (GCNs), it is meaningful to extend our work to cope with these models in the future.

\textbf{Formal privacy guarantee.}
Although we have confirmed that the sparsity-based representation cannot be inverted and thus ensures user privacy, it would be better to provide a formal privacy guarantee. For example, the sparsity-based representation can be further added to some noises based on \textit{differential privacy} \cite{dwork2008differential}, a commonly used technique to avoid privacy leakage. In our future work, we will look for ways to introduce such protection.

\textbf{Real-world applications.}
In this paper, we only evaluate the performance of PFA on simulated datasets and clients. The reason is that it is infeasible to find so many client users and encourage them to conduct our pipeline. However, our simulation design is based on real-world scenarios, which, to some extent, can validate the usefulness of the proposed approach for practical environments.

\section{Related work}

\textbf{Federated Learning and Personalization.}
Federated learning (FL) enables deep learning models to learn from decentralized data without compromising privacy \cite{mcmahan2017communication,kairouz2019advances}. Various research directions have been explored to investigate this promising field, such as  security analysis for FL \cite{melis2019exploiting,bhagoji2019analyzing,bagdasaryan2020backdoor}, efficient communication for FL \cite{konevcny2016federated,alistarh2017qsgd,ivkin2019communication}, personalization for FL \cite{sim2019investigation,jiang2019improving,mansour2020three}, etc.  

This paper focuses on personalization, which stems from the different data distributions of federated clients. There have been a large number of personalization techniques targeting FL. Mansour \textit{et al.}  \cite{mansour2020three} proposed an idea of \textit{user clustering}, where similar clients are grouped together and a separate model is trained for each group. However, this method needs the raw data of each user to implement clustering, which is infeasible due to privacy concerns. In order to guarantee privacy,  Wang \textit{et al.}  \cite{wang2019federated} proposed to utilize transfer learning to achieve personalization, where some or all parameters of a trained federated model are re-learned (i.e., fine-tuned)  on the local data, without any exchange to the user data. Similarly, Jiang \textit{et al.} \cite{jiang2019improving} also conducted fine-tuning to personalize the federated model but this model was generated by a meta-learning \cite{finn2017model} way. Yu \textit{et al.}  \cite{wang2019federated} further extended prior work and systematically evaluated the performance of three personalization methods (i.e., fine-tuning, multi-task learning, knowledge distillation). Different from the above approaches that are implemented in a local client, our framework manages to utilize more useful knowledge existed in other clients by \textit{federated adaptation}, getting rid of the overfitting problem or training bias during the personalization process. Meanwhile, we do not upload any raw data to the server, ensuring user privacy.

\textbf{Sparsity of CNNs.}
Researchers have proposed to take advantage of sparsity in the activation maps to speed up CNNs \cite{graham2017submanifold,parashar2017scnn,shi2017speeding,judd2017cnvlutin2,liu2019wealthadapt,liu2021transtailor}. The main observation is that the Rectified linear unit (ReLU) activation often contains more than 50\% zeros on average. Inspired by this, both hardware-based and software-based convolution algorithms are proposed to exploit the input sparsity. In addition, there are some methods aimed at predicting sparsity in order to skip computation of those unimportant activation spaces \cite{dong2017more,figurnov2017spatially,ren2018sbnet,akhlaghi2018snapea,song2018prediction}. In our work, we attempt to use sparsity to distinguish clients with diverse distributions, a completely different utilization to existing work targeting at acceleration.

\textbf{Privacy Protection of User Data.}
As people are paying more and more attention to their sensitive data, policies such as the General Data Protection Regulation (GDPR) \cite{voigt2017eu} and Health Insurance Portability and Accountability Act (HIPAA) \cite{annas2003hipaa} have been proposed to formally guarantee user privacy. From the technical view, data can be protected by the following two approaches. On one hand, we can rely on secure multiparty computation (SMC) to jointly compute a public function without mutually revealing private inputs by executing cryptographic protocols \cite{naehrig2011can,damgaard2012multiparty,primault2019privacy}. On the other hand, differential privacy (DP) \cite{dwork2008differential,abadi2016deep,geyer2017differentially} can be adopted by adding noises to the data, with the purpose of obfuscating the privacy properties and avoiding the user attributes to be inferred \cite{shokri2017membership,liu2020pmc}. However, SMC requires massive computation power and communication costs, which is unacceptable to the client end. For DP, our sparsity-based representation can be easily added to noises to satisfy the principle of DP, and we leave this implementation in the future work. In addition, Meurisch \textit{et al.} \cite{meurisch2020privacy} presented a new decentralized and privacy-by-design platform, in which different approaches to privacy were adopted to benefit personalization. We believe our approach can also be integrated into the platform to further facilitate the personalization research.

\section{Conclusion}
In this paper, we propose the idea of federated adaptation, which extends existing personalization techniques restricted in a single device. We design and implement a framework named PFA to accomplish federated adaptation in a privacy-preserving manner. PFA takes advantage of the sparsity vector to compare and schedule clients, in order to identify suitable federated clients to conduct the adaptation. Experiments on our simulated datasets and clients demonstrate the effectiveness of PFA, outperforming other personalization methods while ensuring privacy. To the best of our knowledge, this is the first work to study and achieve federated adaptation. We hope our work can provide a new angle to personalization-related research in the FL community.

\begin{acks}
We would like to thank the anonymous reviewers for their valuable feedback.  This work was partly supported by the National Key Research and Development Program (2016YFB1000105) and the National Natural Science Foundation of China (61772042).
\end{acks}

\balance
\bibliographystyle{ACM-Reference-Format}
\bibliography{sample-base}
\clearpage

\end{document}